\pdfoutput=1
\documentclass[conference]{IEEEtran}
\usepackage{cite}
\usepackage{amsmath,amssymb,mathtools,mathrsfs}
\usepackage{graphicx}
\usepackage{booktabs}
\usepackage{array}
\usepackage{multirow}
\usepackage{url}
\usepackage{placeins}

\newcommand{\cmdirDisplaySpacing}{%
  \setlength{\abovedisplayskip}{6pt plus 1pt minus 1pt}%
  \setlength{\belowdisplayskip}{6pt plus 1pt minus 1pt}%
  \setlength{\abovedisplayshortskip}{3pt plus 1pt minus 1pt}%
  \setlength{\belowdisplayshortskip}{6pt plus 1pt minus 1pt}%
  \setlength{\jot}{2pt}}
\AtBeginDocument{\cmdirDisplaySpacing}

\title{Continuous Manifold-Decomposed Impedance\\[-2pt]
Retargeting for Contact-Rich Imitation Learning}

\author{
\IEEEauthorblockN{Liu Jiahao, Kento Kawaharazuka, Tasuku Makabe, Kei Okada}
\IEEEauthorblockA{
\textit{Department of Mechano-Informatics} \\
\textit{Graduate School of Information Science and Technology, The University of Tokyo} \\
Tokyo, Japan \\
\{liu, kawaharazuka, makabe, k-okada\}@jsk.imi.i.u-tokyo.ac.jp
}
}

\begin{document}

\IEEEoverridecommandlockouts

\IEEEaftertitletext{%
\vspace{-6pt}
\begin{center}
  \includegraphics[width=0.96\textwidth]{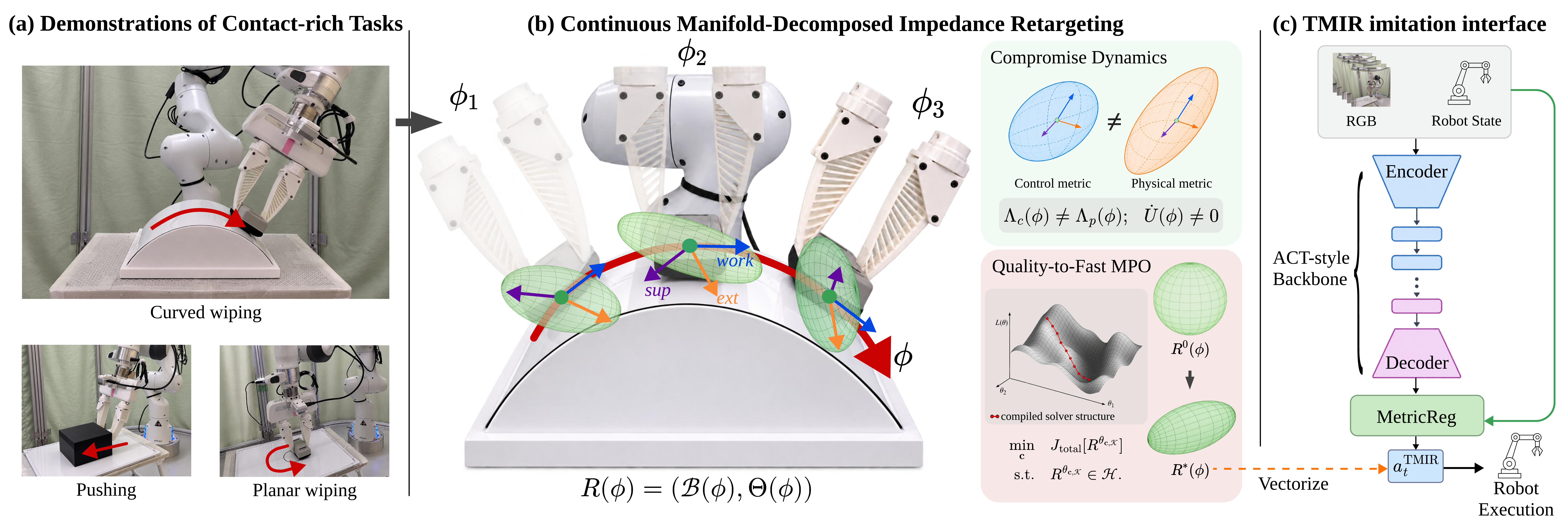}
  \refstepcounter{figure}%
  \label{fig:pipeline}
  \vspace{2pt}
  \parbox{0.96\textwidth}{\footnotesize Fig.~\thefigure. CMDIR lifts pointwise TMIR into a continuous controller field, solves its retargeting through Quality-to-Fast manifold-constrained parameter optimization, and supplies structured variable-impedance supervision for contact-rich imitation.}
\end{center}
\vspace{-4pt}
}

\maketitle

\begin{abstract}
CMDIR extends Manifold-Decomposed Impedance Retargeting (MDIR) to transform fixed-impedance demonstrations into continuous variable-impedance controllers, which can also serve as structured supervision for imitation learning. Continuous Task-Manifold Impedance Representation (TMIR) pairs an evolving task frame with controller instructions. Demo-relative Compromise dynamics retain moving-basis transport and control/physical metric mismatch, yielding displacement, reaction-impulse, and perturbation-sensitivity criteria. Quality-to-Fast automatically compiles a solver structure from development paths within a predefined finite space, re-instantiates that structure for each demonstration, and certifies the resulting candidate by multi-resolution evaluation. Across 225 retargeted-controller trials in three real contact tasks, full CMDIR improves mean task-proxy retention and reduces mean pose deviation, force fluctuation, and peak force relative to discrete MDIR. FastMPO achieves a $5.8$--$9.4\times$ speedup over C-MPO with comparable closed-loop outcomes. Downstream experiments demonstrate learnability of the complete TMIR supervision interface; lower force fluctuation and peak force are observed among successful executions, while completion reliability remains uneven across tasks and environments.
\end{abstract}


\section{Introduction}
Contact-rich manipulation must advance the task and maintain contact while limiting excessive responses. High impedance can amplify contact transients; uniformly reducing gains can weaken progression and support. MDIR addressed this trade-off through task-channel controller retargeting, demonstrating better task preservation than uniform gain scaling \cite{liu2026mdir}.

MDIR pairs semantic task frames with controller instructions in pointwise Task-Manifold Impedance Representations (TMIRs). Extending pointwise retargeting to a continuous controller field requires accounting for basis transport, parameter rates, and overlapping responses between samples; such a field can then be queried at policy rate for imitation supervision. Moving frames and control/physical metric mismatch couple these responses even in orthogonal task coordinates.

CMDIR constructs a continuous field $R(\phi)=(\mathcal B(\phi),\Theta(\phi))$, queryable at policy timestamps up to the source sampling rate. Compromise dynamics balance model detail and source evidence: they retain computable inertia and moving-frame effects, while Demo-relative differences about the same source trajectory cancel common zeroth-order terms. Displacement and impulse bounds restrict modeled task changes; perturbation sensitivity guides moderation. Continuous manifold-constrained parameter optimization (C-MPO) then solves for the continuous retargeted controller field. Quality-to-Fast compiles a reusable solver structure from high-budget development paths within a frozen finite structure space and re-instantiates its numerical updates for each new demonstration (Fig.~\ref{fig:pipeline}).

Our two contributions are: (i) a continuous retargeting formulation that combines TMIR, Compromise dynamics, and displacement, reaction-impulse, and perturbation-sensitivity criteria; and (ii) a Quality-to-Fast numerical scheme that compiles solver structure from development paths within a frozen finite search space, followed by task-specific re-instantiation and multi-resolution post-solve certification.

We evaluate controller retargeting on real planar wiping, curved wiping, and box pushing, and action-chunking-based TMIR supervision through real planar and simulated curved wiping.

\section{Related Work}
\subsection{Variable-Impedance Control and Learning}
Impedance control couples motion and force \cite{hogan1985impedance}. Variants use reinforcement learning \cite{buchli2011variable}, force demonstrations \cite{abudakka2018force}, or geometric gain scheduling \cite{seo2024geometric}. Adaptive Compliance Policy estimates approximate compliance for force moderation and tracking \cite{hou2025adaptive}; San-Miguel et al. tune demonstrated controllers offline under safety and performance conditions \cite{sanmiguel2023condition}. Others address compliance action chunks \cite{kamijo2024compact}, stable impedance \cite{jin2025stable}, compliant force-control dynamics \cite{ge2024compliant}, mobile-manipulation impedance \cite{zhou2025contactvic}, and passive gain modulation \cite{yang2025passivity}. CMDIR retargets fixed-impedance executions through continuous TMIR, source-relative response constraints, and Quality-to-Fast optimization, for policy supervision.

\subsection{Force-Aware Contact-Rich Imitation}
Early computer-controlled bilateral manipulation incorporated tactile and force feedback \cite{inoue1971computer}. Recent systems combine force, tactile, and motion signals in demonstrations and policies \cite{buamanee2024biact,liu2025forcemimic,yu2025mimictouch}. CMDIR uses measured interaction to retarget the demonstrated controller, then supplies its optimized TMIR as structured policy supervision.

\subsection{Demonstration Transformation and Controller Retargeting}
Demonstration transformation includes geometric or visual generation, embodiment transfer, and physics-based synthesis \cite{mandlekar2023mimicgen,xue2025demogen,yang2025physicsgen,li2025okami,liu2025immimic}. MDIR instead operates on the executed controller through pointwise TMIR objects and task-relevant response retention \cite{liu2026mdir}. CMDIR builds on this controller-retargeting foundation by defining the representation and response criteria on a shared continuous field.

The resulting per-demonstration optimization connects to work on accelerating repeated constrained problems through learned warm starts \cite{sambharya2024warmstart} and active-set prediction with full-problem fallback \cite{dyrska2024active}. Quality-to-Fast compiles reusable subproblem order, near-active constraint families, and repair sequences from development paths, while recomputing controller values and responses for each demonstration.

\section{Continuous Controller Retargeting}
\subsection{From Source Execution to Continuous TMIR}
\label{sec:formulation}

Let $\mathcal D$ contain measured pose $T_{\rm real}$, command pose $T_{\rm cmd}$, velocity $\dot x_{\rm real}$, wrench $F_{\rm sensor}$, fixed gains $K_{\rm src},D_{\rm src}$ over $[t_0,t_T]$. Controller fields use normalized time $\phi(t)=(t-t_0)/(t_T-t_0)\in[0,1]$; derivatives and response integrals retain physical time. Source quantities also use the known robot model and, where needed, source-scene geometry. For offline contact-state segmentation, a frozen temporal convolutional network (TCN) \cite{lea2017tcn}, trained on ten manually labeled executions per task, assigns $\ell_{\rm c}(t)\in\{\mathrm{Free},\mathrm{Stable},\mathrm{Unstable}\}$: Stable denotes steady wiping or pushing, and Unstable denotes collision. Labels fix evaluation windows without changing frame geometry.

Two positive-definite metrics distinguish controller geometry from physical response. Following MDIR \cite{liu2026mdir}, the source-defined control metric $\Lambda_c(t)$ defines task-coordinate normalization and orthogonality; Cartesian operational-space inertia $\Lambda_p(t)$ weights physical response. Both are evaluated along the source trajectory and frozen with respect to candidate search, together with contact labels and response scales.

Discrete MDIR defines $R_k=(\mathcal B_k,\Theta_k)$ at sample $k$, with the ordered Work--Exertion--Support frame $\mathcal B_k=(u_{W,k},u_{E,k},u_{S,k})$ and instructions $\Theta_k=\{(k_{i,k},d_{i,k},\delta_{i,k})\}_{i\in\mathcal I_k}$. CMDIR defines the continuous counterpart over normalized time $\phi$:
\begingroup
\begin{equation}
\begin{aligned}
R(\phi)&=(\mathcal B(\phi),\Theta(\phi)),\\
\mathcal B(\phi)&=(u_W(\phi),u_E(\phi),u_S(\phi)),\\
\Theta(\phi)&=\{(k_i(\phi),d_i(\phi),\delta_i(\phi))\}_{i\in\mathcal I(\phi)}.
\end{aligned}
\label{eq:tmir-continuous}
\end{equation}
\endgroup
The active-channel sets $\mathcal I_k,\mathcal I(\phi)$ pair each direction with stiffness $k_i$, damping $d_i$, and offset $\delta_i$ from the demonstrated command pose. Passive impedance acts on the metric complement of $\mathcal M_{\rm task}=\operatorname{span}(U_{\rm task})$.

At every phase query, $\mathcal B(\phi)$ is represented by the ordered metric-orthonormal frame
\begin{equation}
\begin{aligned}
U_{\rm task}(\phi)&=[u_W(\phi),u_E(\phi),u_S(\phi)]\in\mathbb R^{6\times3},\\
U_{\rm task}^\mathsf{T}\Lambda_cU_{\rm task}&=I_3,
\qquad W_{\rm task}=\Lambda_cU_{\rm task} .
\end{aligned}
\label{eq:frame}
\end{equation}
Cartesian twist $\dot x$ and wrench $F$ map to $\dot s=W_{\rm task}^{\mathsf T}\dot x$ and $Q=U_{\rm task}^{\mathsf T}F$. Similarly, $e=W_{\rm task}^{\mathsf T}e_x$, where $e_x$ is the demonstrated command-to-measured-state pose error in the same six-dimensional convention.

Construction follows $u_W^{\rm acc}\longrightarrow u_E^{\rm acc}\longrightarrow u_S$. Work follows accepted motion; Ext (Exertion) follows interaction evidence and force memory after removing Work; Support supplies the remaining support direction. Nowork denotes joint Ext--Support. Hemisphere alignment and boundary-matched continuation preserve sign and derivative continuity, including dependence on $\Lambda_c(t)$.

Fission continuously establishes Ext between Pure Support and mature joint Ext--Support. Its weight $\gamma(\tau)=6\tau^5-15\tau^4+10\tau^3$, $\tau\in[0,1]$, has zero endpoint slope and curvature. All axes remain extended while weights change; each response window belongs to one Pure Support, Fission, or Mature family.

The same basis analytically compiles the Cartesian-to-Manifold (C2M) reference $R^0=(\mathcal B,\Theta^0)$, where $\Theta^0=\{(k_i^0,d_i^0,\delta_i^0)\}_{i\in\mathcal I}$. Its TMIR reconstruction gives $(K^0,D^0,T_{\rm cmd}^0)$ and commands $Q_i^0$, preserving $Q_{\rm ctrl}^0(z_{\rm demo})=Q_{\rm ctrl}^{\rm demo}(z_{\rm demo})$ at the source state. The demonstration's full task-coordinate feedback matrices remain the response reference.

\subsection{Continuous Candidate TMIR Field}

With $\mathcal B(\phi)$ fixed by the source, optimization changes the executable instruction $\Theta(\phi)$. Seven nonuniform cubic B-spline fields parameterize these changes:
\begin{equation}
\theta_{\mathbf c,\mathcal K}(\phi)
=\mathscr S_{\mathcal K}(\mathbf c)(\phi)=\theta(\phi).
\label{eq:theta}
\end{equation}
Here $\theta=(\alpha_W,\alpha_E,\alpha_S,\zeta_W,\delta_W^m,\delta_E^m,\delta_S^m)$, $\mathbf c$ contains the spline coefficients, and $\mathcal K$ is the knot vector. Thus $\theta$ parameterizes the executable instruction $\Theta^\theta$. With $k_i^{\rm floor}=\min(\chi_i k_{{\rm pass}i},k_i^0)$, stiffness and damping are
\begingroup\small\cmdirDisplaySpacing
\begin{equation}
\begin{aligned}
k_i^\theta&=k_i^{\rm floor}+\alpha_i(k_i^0-k_i^{\rm floor}),\\
d_W^\theta&=2\zeta_W\sqrt{k_W^\theta},\\
d_i^\theta&=d_i^0
\sqrt{\frac{k_i^\theta}{k_i^0}},
\qquad i\in\{E,S\}.
\end{aligned}
\label{eq:impedance}
\end{equation}
\endgroup
with $0\leq\alpha_i\leq1$. The source-fixed $\chi_i$ records support for channel $i$, and $k_{{\rm pass}i}$ is its passive background stiffness. The associated command is
\begin{equation}
Q_i^\theta=k_i^\theta(e_i+\delta_i^\theta)
-d_i^\theta\dot s_i .
\label{eq:command}
\end{equation}
The C2M offset $\delta_i^0$ and applied correction $\delta_i^m$ give the executable offset $\delta_i^\theta=\delta_i^0+\delta_i^m$; the spring displacement is $e_i+\delta_i^\theta$. These offsets enter $\Theta^\theta$ and reconstruct $K^\theta,D^\theta,T_{\rm cmd}^\theta$ with the passive complement. Setting $\alpha_i=1$, $\delta_i^m=0$, and $\zeta_W=d_W^0/(2\sqrt{k_W^0})$ recovers $R^0$, including its command.

C2M fixes the parameterization origin, scales, and boxes; Demo supplies the task-response reference.

\subsection{Demo-Relative Compromise Dynamics and Retained Responses}

To evaluate a candidate's task effect, let $\Delta_\theta f=f^\theta-f^{\rm demo}$. Although $U_{\rm task}^{\mathsf T}\Lambda_cU_{\rm task}=I$, physical inertia need not share the control metric, and the frame changes along the trajectory. Retaining these effects gives the first-order Compromise dynamics

\begin{equation}
\begin{aligned}
M_0\Delta_\theta\ddot s+C_0\Delta_\theta\dot s
&=\Delta_\theta Q_{\rm ctrl}+\Delta_\theta Q_{\rm env},\\
M_0&=U_{\rm task}^\mathsf{T}\Lambda_pU_{\rm task},
\qquad C_0=U_{\rm task}^\mathsf{T}\Lambda_p\dot U_{\rm task} .
\end{aligned}
\label{eq:relative}
\end{equation}
Candidate and Demo are expanded about the same source trajectory, so their common zeroth-order terms cancel in the relative response. We retain the source-computable first-order effects of projected physical inertia $M_0(t)$ and control-metric frame transport $C_0(t)$. These account for metric mismatch and moving-basis coupling while preserving role-wise controller parameters. The Ideal comparator sets $M_0=I_m,C_0=0$, with full propagation rank $m=3$. Other time-varying operational-space inertial and implementation residuals, denoted $F_{\rm op,res}$, remain outside this model.

\begin{figure*}[!t]
  \centering
  \includegraphics[width=\textwidth]{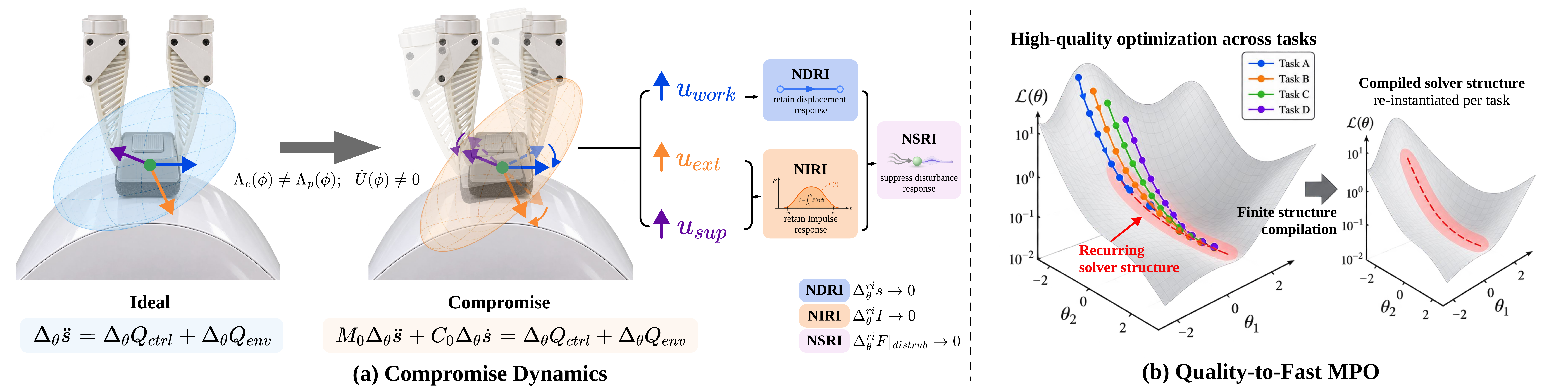}
 \caption{Schematic illustration of Compromise dynamics and Quality-to-Fast numerical design. (a) Compromise dynamics propagates retained responses with projected physical inertia and moving-frame transport. (b) Quality-to-Fast abstracts accepted development paths into solver-structure signatures, compiles a reusable structure within a finite search space, and recomputes all numerical updates for each new demonstration. The paths are illustrative; compilation uses three development demonstrations.}
  \label{fig:dynamics-q2f}
\end{figure*}

These dynamics define three Relative Inducers (RIs) on physical-time windows $\omega=[t_s,t_s+T]$. For a retained submanifold $\mathcal M_{\mathcal A}$ of rank $r$, source-fixed $S_{\mathcal A}\in\mathbb R^{r\times m}$ selects open-response coordinates, while $A_{\mathcal A}\in\mathbb R^{r\times m}$ constrains relative velocity and induces reaction $A_{\mathcal A}^{\mathsf T}\lambda_{\mathcal A}$. The Nominal Displacement Relative Inducer (NDRI) and Nominal Impulse Relative Inducer (NIRI) bound normalized displacement and constraint-reaction impulse changes. The Nominal Sensitivity Relative Inducer (NSRI) evaluates perturbation-induced feedback action for the objective.

In constructing the task manifold, Work carries task motion, while Ext and Support form the nonmoving Nowork subspace, with $\dot s_N=0$. We therefore apply NDRI to preserve Work displacement response and joint NIRI to preserve the reaction impulse required to maintain the Nowork constraint.

\paragraph{NDRI: task displacement}
To retain task progression, consider the open response with $\Delta_\theta Q_{\rm env}=0$. Let $\mathcal G_D$ be the zero-initial-condition Green operator induced by $M_0$ and $C_0$. The retained terminal displacement is
\begin{equation}
\begin{aligned}
\mathcal R_{D,\mathcal A}^{T}[Q_{\mathcal A}]
&=S_{\mathcal A}(t_s+T)
\int_{t_s}^{t_s+T}\!
\mathcal G_D(t_s+T,\tau)\\
&\quad\cdot S_{\mathcal A}(\tau)^\mathsf{T}
Q_{\mathcal A}(\tau)\,d\tau ,
\end{aligned}
\label{eq:ndri-response}
\end{equation}
where $Q_{\mathcal A}=S_{\mathcal A}Q_{\rm ctrl}$. Work displacement retention uses
\begingroup\small\cmdirDisplaySpacing
\begin{equation}
\begin{aligned}
g_{\rm NDRI}(\omega)
&=\frac{\|\Delta_\theta\mathcal R_{D,\mathcal A_W}^{T_D}(\omega)\|_{G_{D,\mathcal A_W}}}
{S_{D,\mathcal A_W}^{\rm demo}(\omega)}-\tau_D\\
&\leq0,
\qquad \omega\in\mathcal W_D.
\end{aligned}
\label{eq:ndri}
\end{equation}
\endgroup
Normalized Work coordinates give $G_{D,\mathcal A_W}=1$. The scale is the larger of the demonstrated terminal-response magnitude and $\tfrac12 T_D^2\operatorname{RMS}(Q_W^{\rm demo})$ over Work-supported source times, avoiding a vanishing reference through cancellation. Allowance $\tau_D$ applies to complete Work windows $\mathcal W_D$ in all three contact states.

\paragraph{NIRI: constrained impulse}
Contact retention also depends on the reaction required to maintain constrained motion. Candidate and Demo share $A_{\mathcal A}\Delta_\theta\dot s=0$, hence
\begin{equation}
A_{\mathcal A}\Delta_\theta\ddot s+
\dot A_{\mathcal A}\Delta_\theta\dot s=0 .
\label{eq:constraint}
\end{equation}
Let $H_{\mathcal A}=A_{\mathcal A}M_0^{-1}A_{\mathcal A}^{\mathsf T}$. The constraint-reaction increment is
\begin{equation}
\Delta_\theta\lambda_{\mathcal A}
=-H_{\mathcal A}^{-1}\!
\left[
\dot A_{\mathcal A}\Delta_\theta\dot s+
A_{\mathcal A}M_0^{-1}
(\Delta_\theta Q_{\rm ctrl}^{\rm nom}-C_0\Delta_\theta\dot s)
\right],
\label{eq:niri-lambda}
\end{equation}
The retained impulse change is $\Delta I_{\mathcal A}^\theta(\omega)=\int_\omega\Delta_\theta\lambda_{\mathcal A}(t)\,dt$. Pure Support uses
$A_{\rm PS}=[0\ 0\ 1]$; Fission and Mature use
$A_N=[\,0\ 1\ 0;\ 0\ 0\ 1\,]$. On complete, family-consistent Stable windows,
\begingroup\small\cmdirDisplaySpacing
\begin{equation}
\begin{aligned}
g_{{\rm NIRI},\mathcal F}(\omega)
&=\frac{\|\Delta I_{\mathcal F}^\theta(\omega)\|_{G_{I,\mathcal F}^{-1}}}
{S_{I,\mathcal F}^{\rm task}}-\tau_I\\
&\leq0,\qquad \omega\in\mathcal W_{I,\mathcal F}.
\end{aligned}
\label{eq:niri}
\end{equation}
\endgroup
Each family $\mathcal F$ uses normalized coordinates, $G_{I,\mathcal F}=I_r$. Its fixed impulse scale is the window duration times the RMS norm of the demonstrated generalized action over that family's complete-window support. Allowance $\tau_I$ bounds impulse changes on Stable windows $\mathcal W_{I,\mathcal F}$; Unstable events enter the objective below.

\paragraph{NSRI: perturbation sensitivity}
Beyond nominal task retention, NSRI measures how strongly the controller reacts to a local state perturbation. With task-coordinate matrices $K_s^\theta$ and $D_s^\theta$, the equilibrium perturbation follows
\begin{equation}
M_0\ddot{\widetilde s}+(D_s^\theta+C_0)\dot{\widetilde s}
+K_s^\theta\widetilde s=\widetilde Q_{\rm env}.
\label{eq:nsri-dynamics}
\end{equation}
The feedback perturbation is $\widetilde Q_{\rm ctrl}^\theta=-K_s^\theta\widetilde s-D_s^\theta\dot{\widetilde s}$; $K_s^\theta,D_s^\theta$ are obtained from the reconstructed controller in task coordinates. NSRI propagates an initial state perturbation with $\widetilde Q_{\rm env}=0$. The output for $\mathcal M_{\mathcal A}$ is
$C_{\mathcal A}^\theta=S_{\mathcal A}[K_s^\theta\ D_s^\theta]$. With the closed-loop transition $\Phi_{\rm cl}^\theta$, the finite-horizon output-energy matrix is
\begin{equation}
\mathcal W_{\mathcal A}^\theta(t_s)=
\int_{t_s}^{t_s+T}\!
\Phi_{\rm cl}^\theta(t,t_s)^\mathsf{T}
C_{\mathcal A}^{\theta\mathsf{T}}(t)C_{\mathcal A}^{\theta}(t)
\Phi_{\rm cl}^\theta(t,t_s)\,dt .
\label{eq:nsri-gramian}
\end{equation}
The matrix $E_{\mathcal A}$ embeds a rank-$r$ perturbation into the full position--velocity state; $\mathcal S_{T,\mathcal A}=\operatorname{diag}(I_r,T I_r)$ puts velocity on the displacement scale. They give
\begin{equation}
P_{\mathcal A}^\theta(\omega)=\sigma_{\mathcal A}^2(\omega)
\lambda_{\max}\!\left[
\mathcal S_{T,\mathcal A}^{-\mathsf T}E_{\mathcal A}^{\mathsf T}
\frac{\mathcal W_{\mathcal A}^\theta}{T}
E_{\mathcal A}\mathcal S_{T,\mathcal A}^{-1}
\right].
\label{eq:nsri-power}
\end{equation}
The within-window variance of demonstrated pose error projected onto the retained axis, $\sigma_{\mathcal A}^2$, fixes perturbation magnitude. On the same windows,
\begin{equation}
J_{{\rm NSRI},\mathcal A}=
\frac{\operatorname{Mean}_\omega P_{\mathcal A}^\theta}
{\operatorname{Mean}_\omega P_{\mathcal A}^{\rm demo}}
+\beta_{\mathcal A}
\frac{\operatorname{CVaR}_{95,\omega}(P_{\mathcal A}^\theta)}
{\operatorname{CVaR}_{95,\omega}(P_{\mathcal A}^{\rm demo})}.
\label{eq:nsri}
\end{equation}
The weight $\beta_{\mathcal A}$ balances mean and upper-tail sensitivity. CMDIR evaluates NSRI on rank-one Work, Ext, and Support over Stable windows, using it to moderate perturbation response within the NDRI/NIRI task-retention boundary.

\subsection{Continuous Manifold-Constrained Optimization}

Continuous manifold-constrained parameter optimization (C-MPO) minimizes
 $J_{\rm total}=J_K+J_D+J_{\rm cmd\ pose}+J_{\rm stable}+J_{\rm danger}$.
$J_K$ combines time means of squared stiffness fractions $\alpha_i$ and scaled stiffness phase derivatives; $J_D$ combines squared scaled damping magnitudes and phase derivatives. Separate gradient-based scales computed before optimization balance magnitude and rate equally. $J_{\rm cmd\ pose}$ aggregates squared positive command-speed excess over Demo by its time mean and fixed-duration upper tail.

For $i\in\{W,E,S\}$, local command variation and perturbation sensitivity form the role-wise stability term:
\begingroup\small\cmdirDisplaySpacing
\begin{equation}
\begin{aligned}
V_{Q,i}^\theta(\omega)&=\frac{1}{|\omega|}\int_\omega
\left(\frac{Q_i^\theta(t)-\overline Q_i^\theta(\omega)}{s_{Q,i}}\right)^2dt,\\
J_{{\rm stable},i}&=\frac12\frac{J_{Q\text{-var},i}}
{N_{Q,i}^{\rm src}}+
\frac12\frac{J_{{\rm NSRI},i}}
{N_{{\rm NSRI},i}^{\rm src}}.
\end{aligned}
\label{eq:stable}
\end{equation}
\endgroup
Here $J_{\rm stable}=\sum_iJ_{{\rm stable},i}$, $s_{Q,i}$ scales channel commands, and $J_{Q\text{-var},i}$ aggregates their variance by mean and CVaR over Stable windows. The source-derived $N^{\rm src}$ factors normalize command variation and perturbation sensitivity before optimization.

For an Unstable event $e$ on interval $\mathcal U_e$, local point clouds determine the obstacle normal $n_{G,e}$; source force selects its inward sign and contact support. Define $h_{G,e}^{\mathsf T}=[n_{G,e}^{\mathsf T}\ 0_{1\times3}]$ and $F_{\rm exerted}^{\rm demo}=-F_{\rm sensor}^{\rm demo}$. The inward force $r_{G,e}^{\rm demo}=[h_{G,e}^{\mathsf T}F_{\rm exerted}^{\rm demo}]_+$ gives event impulse $I_e^{\rm demo}$ and root-mean-square force $F_{e,{\rm rms}}^{\rm demo}$. Let
$A_{G,e}=h_{G,e}^{\mathsf T}U_{\rm active}$, where
$U_{\rm active}=[u_W,u_S]$ for Pure Support and
$U_{\rm active}=[u_W,u_E,u_S]$ otherwise. Using $M_0,C_0,\Delta Q$ in this same active domain, \eqref{eq:niri-lambda} yields
\begin{equation}
\begin{aligned}
\widehat r_{G,e}^\theta(t)
&=[r_{G,e}^{\rm demo}(t)-\Delta_\theta\lambda_{G,e}(t)]_+.
\end{aligned}
\label{eq:danger-response}
\end{equation}
Its event integral $I_e^\theta$ and RMS value $F_{e,{\rm rms}}^\theta$ define
\begin{equation}
J_{\rm danger}=
\sum_e\frac12\left[
\left(\frac{I_e^\theta}{I_e^{\rm demo}}\right)^2+
\left(\frac{F_{e,{\rm rms}}^\theta}{F_{e,{\rm rms}}^{\rm demo}}\right)^2
\right].
\label{eq:danger}
\end{equation}
This objective moderates the modeled event reaction relative to Demo. It evaluates a controller-response proxy, rather than predicting sensor force.

Work damping is bounded using a normalized coordinate-energy budget. For each source-supported $\omega=[t_s,t_s+T_E]$, let
$P_W^+=[Q_{{\rm env},W}^{\rm demo}\dot s_W-F_{\rm noise}|\dot s_W|]_+$. The Work damping bound is
\begingroup\small\cmdirDisplaySpacing
\begin{equation}
\begin{aligned}
E_W^{\rm src}(\omega)
&=\tfrac12\dot s_W^2(t_s)+\int_\omega P_W^+(t)\,dt,\\
C_W^\theta(\omega)
&=\int_{\omega}2\sqrt{k_W^\theta(t)}\,\dot s_W^2(t)\,dt,\\
\zeta_{W,{\rm lb}}(t)
&=(1-\chi_{\rm kin})
\frac{d_{{\rm passi},W}}{2\sqrt{k_W^\theta}}
+\chi_{\rm kin}\max_{\omega\ni t}\frac{E_W^{\rm src}(\omega)}{C_W^\theta(\omega)},\\
&\hspace{16mm}\zeta_W^\theta(t)\geq\zeta_{W,{\rm lb}}(t)\geq0.
\end{aligned}
\label{eq:damping-energy}
\end{equation}
\endgroup
Here $T_E=0.20\,{\rm s}$, $d_{{\rm passi},W}$ is projected passive damping, and $\chi_{\rm kin}\in[0,1]$ is the source-derived kinetic-evidence gate. Energy demand $E_W^{\rm src}$ and unit-ratio dissipation capacity $C_W^\theta$ set a normalized-coordinate dissipation bound without a prescribed decay rate. This budget differs from physical kinetic energy under $M_0$ and does not establish passivity.

Together these constraints define $\mathcal H=\mathcal H_{\rm NDRI}^{W}\cap\mathcal H_{\rm NIRI}^{N}\cap\mathcal H_{\zeta_W}\cap\mathcal H_\alpha$, where $\mathcal H_\alpha$ enforces the parameter boxes. Each component is $\mathcal H_j=\{R:g_j[R;\xi]\leq0,\ \forall\xi\in\mathbb X_j\}$ on its prescribed query domain. For source instance $\mathfrak p$, C-MPO solves
\begin{equation}
\begin{aligned}
\min_{\mathbf c}\quad&J_{\rm total}[R^{\theta_{\mathbf c,\mathcal K}}]\\
\mathrm{s.t.}\quad&
R^{\theta_{\mathbf c,\mathcal K}}\in\mathcal H.
\end{aligned}
\label{eq:cmpo}
\end{equation}

C-MPO proposes coefficient updates using the augmented Lagrangian
\begingroup\small\cmdirDisplaySpacing
\begin{equation}
\begin{aligned}
\mathcal L_\kappa(\mathbf c,\boldsymbol\mu)
&=J_{\rm total}[R^{\theta_{\mathbf c,\mathcal K}}]
+\sum_j\Psi_{\kappa,j}(\mathbf c,\boldsymbol\mu_j),\\
\Psi_{\kappa,j}
&=\max_{\xi\in\mathbb X_{j,k}^{\rm iter}}
\frac{[\mu_{j,\xi}+\kappa g_j(R^{\theta_{\mathbf c,\mathcal K}};\xi)]_+^2
-\mu_{j,\xi}^2}{2\kappa}.
\end{aligned}
\label{eq:alpotential}
\end{equation}
\endgroup
Here $k$ indexes iterations, $\mu_{j,\xi}\ge0$ are multipliers, and $\kappa>0$ controls the penalty. Only feasible, objective-decreasing updates are retained. After fixed-space convergence, positive independent-validation residuals trigger local refinement $\mathcal K^+=\mathcal K\cup\mathcal K_{\rm viol}$ with function-preserving insertion, $R^{\theta_{\mathbf c^+,\mathcal K^+}}=R^{\theta_{\mathbf c,\mathcal K}}$. Final candidates undergo multi-resolution evaluation.

\section{Quality-to-Fast Optimization}
C-MPO directly searches coefficients, Quality performs higher-budget offline exploration, and FastMPO executes compiled solver structure; all solve the same CMDIR problem. For instance $\mathfrak p$, write $J_{\mathfrak p}(\theta)=J_{\rm total}[R^\theta]$. Phase-I supplies a validated initialization $\theta_{0,\mathfrak p}$; $\mathcal H_{\mathfrak p}^{\Sigma}$ denotes feasibility under the shared source, independent-validation, and dense queries.

\subsection{Quality Paths}
Development proceeds through
\begin{equation}
\begin{aligned}
\theta_{0,\mathfrak p}&\xrightarrow{\mathrm{C\text{-}MPO}}\theta_{C,\mathfrak p},
\qquad \operatorname{Q0}_{\mathfrak p}(\theta_{C,\mathfrak p})=1,\\
\theta_{C,\mathfrak p}&\xrightarrow{\mathrm{Quality}}\theta_{Q,\mathfrak p}.
\end{aligned}
\label{eq:quality-path}
\end{equation}
Q0 checks reconstruction, objective, constraints, and problem identity at $\theta_C$. Quality searches predefined semantic subproblems, accepting only certified-feasible transitions that decrease $J_{\mathfrak p}$. The development path $\Pi_{D,\mathfrak p}=\Pi_{C,\mathfrak p}\oplus\Pi_{Q,\mathfrak p}$ concatenates the accepted C-MPO trace $\theta_0\to\theta_C$ and Quality trace $\theta_C\to\theta_Q$. Its signatures $(\rho_k,\mathfrak b_k,\mathfrak r_k,\prec_k)$ record scope, active-constraint family, recovery sequence, and evaluation order. $\operatorname{PathAbs}(\{\Pi_{D,\mathfrak p}\}_{\mathfrak p\in\mathscr P_{\rm dev}})=\mathfrak G_Q$ retains these signatures and discards coefficients, numerical directions, residual magnitudes, task-manifold and C2M values, RI operators, and all task-specific numerical data.

\subsection{Finite Structure Compilation}
Compilation is automatic within a predefined finite structure space. The structure alphabet, shared preconditioner set $\mathfrak P$, maximum length $L_{\max}$, and quality gates are frozen beforehand:
\begingroup\small\cmdirDisplaySpacing
\begin{equation}
\mathfrak C_F=\left\{\mathfrak S(\mathbf s,P):
\begin{array}{l}
|\mathbf s|\le L_{\max},\quad P\in\mathfrak P,\\
\operatorname{supp}(\mathbf s)\ge m_{\min}
\end{array}\right\}.
\label{eq:compiler-space}
\end{equation}
\endgroup
Here $\operatorname{supp}(\mathbf s)$ counts the development instances containing signature sequence $\mathbf s$. Re-instantiation on each instance gives $\theta_{F,\mathfrak p}^{\mathfrak S}$, runtime $T_{\mathfrak p}$, and relative objective gap $\Delta_{\mathfrak p}=(J_{\mathfrak p}(\theta_{F,\mathfrak p}^{\mathfrak S})-J_{R,\mathfrak p})/\max(|J_{R,\mathfrak p}|,\varepsilon)$. For C-MPO-level validation, $J_{R,\mathfrak p}=J_{\mathfrak p}(\theta_{C,\mathfrak p})$; $\varepsilon>0$ protects a zero denominator. Selection is
\begingroup\small\cmdirDisplaySpacing
\begin{equation}
\begin{aligned}
\operatorname{Pass}_{\mathfrak p}(\mathfrak S)
&=\mathbb I\left[
\begin{array}{l}
\theta_{F,\mathfrak p}^{\mathfrak S}\in\mathcal H_{\mathfrak p}^{\Sigma},\\
\Delta_{\mathfrak p}\le\varepsilon_J,\quad T_{\mathfrak p}\le T_{\max}
\end{array}\right],\\
\mathfrak S_F
&=\operatorname*{lex\,arg\,min}_{\substack{\mathfrak S\in\mathfrak C_F\\
\operatorname{Pass}_{\mathfrak p}(\mathfrak S)=1\;\forall\mathfrak p}}
\left(\max_{\mathfrak p}\Delta_{\mathfrak p},
\max_{\mathfrak p}T_{\mathfrak p},|\mathbf s|,
\operatorname{Code}(\mathfrak S)\right).
\end{aligned}
\label{eq:compiler-selection}
\end{equation}
\endgroup
The gates require multi-resolution feasibility, bounded quality loss, and bounded runtime. Selection minimizes worst-case gap, runtime, and length; encoding breaks ties. Compilation fails if none passes.

\subsection{Fast Re-instantiation and Certification}
On a new demonstration $\mathfrak q$,
\begingroup\small\cmdirDisplaySpacing
\begin{equation}
\begin{aligned}
\widehat\theta_{\mathfrak q}
&=\operatorname{FastInst}_{\mathfrak q}(\mathfrak S_F,\theta_{0,\mathfrak q}),\\
\widetilde\theta_{\mathfrak q}
&=\operatorname{Rep}_{B,\mathfrak q}(\widehat\theta_{\mathfrak q},\theta_{0,\mathfrak q}),\\
\theta_{F,\mathfrak q}
&=\begin{cases}
\widetilde\theta_{\mathfrak q},
&\begin{gathered}
\operatorname{Cert}_{\mathfrak q}^{\Sigma}(\widetilde\theta_{\mathfrak q})=1,\\
J_{\mathfrak q}(\widetilde\theta_{\mathfrak q})\le J_{\mathfrak q}(\theta_{0,\mathfrak q}), 
\end{gathered}\\
\theta_{0,\mathfrak q},&\text{otherwise}.
\end{cases}
\end{aligned}
\label{eq:fast-accept}
\end{equation}
\endgroup
FastInst recomputes directions, constraints, RI operators, and updates from the new initialization; bounded recovery restores feasibility. Certification and objective non-degradation determine acceptance. Instances without a validated initialization are rejected. No development $\theta_C$, $\theta_Q$, task-manifold values, active-constraint values, or controller coefficients are transferred. The returned field is $R^\star=R^{\theta_{F,\mathfrak q}}$.

\par\smallskip
\noindent\begin{minipage}{\columnwidth}
\small
\hrule\smallskip
\textbf{Algorithm 1}\quad Quality-to-Fast MPO
\smallskip\hrule\smallskip
\begin{tabular}{@{}r@{\quad}p{0.90\columnwidth}@{}}
1 & For each development instance $\mathfrak p\in\mathscr P_{\rm dev}$:\\
2 & \quad $(\theta_{C,\mathfrak p},\Pi_{C,\mathfrak p})\leftarrow\mathrm{C\text{-}MPO}(\theta_{0,\mathfrak p})$.\\
3 & \quad Require $\operatorname{Q0}_{\mathfrak p}(\theta_{C,\mathfrak p})=1$.\\
4 & \quad $\Pi_{Q,\mathfrak p}\leftarrow\mathrm{Quality}(\theta_{C,\mathfrak p})$.\\
5 & \quad $\Pi_{D,\mathfrak p}\leftarrow\Pi_{C,\mathfrak p}\oplus\Pi_{Q,\mathfrak p}$; end for.\\
6 & $\mathfrak G_Q\leftarrow\operatorname{PathAbs}(\{\Pi_{D,\mathfrak p}\})$; enumerate $\mathfrak S\in\mathfrak C_F$.\\
7 & Re-instantiate each structure on every development instance.\\
8 & Select $\mathfrak S_F$ by (\ref{eq:compiler-selection}); stop if none passes.\\
9 & New $\mathfrak q$: $\widetilde\theta_{\mathfrak q}\leftarrow\operatorname{Rep}_{B,\mathfrak q}(\operatorname{FastInst}_{\mathfrak q}(\mathfrak S_F,\theta_{0,\mathfrak q}),\theta_{0,\mathfrak q})$.\\
10 & Return $\widetilde\theta_{\mathfrak q}$ if certified and $J_{\mathfrak q}(\widetilde\theta_{\mathfrak q})\le J_{\mathfrak q}(\theta_{0,\mathfrak q})$; otherwise return $\theta_{0,\mathfrak q}$.\\
\end{tabular}
\smallskip\hrule
\end{minipage}
\par\smallskip

\begin{table*}[t]
\caption{Controller-retargeting closed-loop evaluation.}
\label{tab:controller}
\centering
\scriptsize
\renewcommand{\arraystretch}{0.90}
\setlength{\tabcolsep}{1.4pt}
\begin{tabular*}{\textwidth}{@{\extracolsep{\fill}}llcccccc@{}}
\toprule
Task & Method & Task proxy & Pose $\Delta$ (\%) & Force fluctuation $\Delta$ (\%) & Peak force $\Delta$ (\%) & Nominal power $\Delta$ (\%) & Solve time (s) \\
\midrule
\multirow{5}{*}{\shortstack{Curved\\wiping}}
& $M_1$ & $0.696\pm0.045$ & $0.617\pm0.077$ & $-3.93\pm20.48$ & $-2.64\pm21.97$ & $-7.05\pm2.28$ & $680.1\pm194.3$ \\
& $M_2$ & $0.378\pm0.095$ & $1.334\pm0.117$ & $+9.16\pm22.05$ & $+71.43\pm79.12$ & $-15.08\pm5.39$ & $250.2\pm55.8$ \\
& $M_3$ & $0.775\pm0.134$ & $0.320\pm0.077$ & $-24.25\pm21.81$ & $-23.37\pm11.16$ & $\mathbf{-19.50\pm2.81}$ & $686.9\pm47.8$ \\
& $M_4$ & $\mathbf{0.869\pm0.070}$ & $\mathbf{0.235\pm0.061}$ & $\mathbf{-26.75\pm24.67}$ & $\mathbf{-30.91\pm22.53}$ & $-17.82\pm3.28$ & $975.8\pm230.0$ \\
& $M_5$ & $0.847\pm0.080$ & $0.245\pm0.063$ & $-25.74\pm25.14$ & $-28.05\pm26.95$ & $-17.98\pm2.97$ & $\mathbf{124.95\pm17.96}$ \\
\cmidrule(lr){2-8}
\multirow{5}{*}{\shortstack{Planar\\wiping}}
& $M_1$ & $0.796\pm0.083$ & $0.681\pm0.147$ & $-3.51\pm10.49$ & $-12.51\pm8.18$ & $-12.64\pm1.67$ & $248.7\pm84.7$ \\
& $M_2$ & $0.472\pm0.045$ & $1.794\pm0.572$ & $+34.75\pm70.95$ & $+31.91\pm68.25$ & $\mathbf{-15.83\pm7.53}$ & $106.9\pm12.3$ \\
& $M_3$ & $0.820\pm0.089$ & $0.581\pm0.218$ & $\mathbf{-16.85\pm13.72}$ & $\mathbf{-22.81\pm17.45}$ & $-15.62\pm2.60$ & $568.4\pm92.1$ \\
& $M_4$ & $0.833\pm0.086$ & $\mathbf{0.495\pm0.175}$ & $-14.30\pm12.57$ & $-20.38\pm15.72$ & $-14.29\pm3.81$ & $657.9\pm141.0$ \\
& $M_5$ & $\mathbf{0.835\pm0.092}$ & $0.506\pm0.143$ & $-14.22\pm13.15$ & $-15.98\pm19.27$ & $-14.21\pm3.38$ & $\mathbf{70.3\pm4.3}$ \\
\cmidrule(lr){2-8}
\multirow{5}{*}{\shortstack{Box\\pushing}}
& $M_1$ & $0.929\pm0.012$ & $1.565\pm0.264$ & $-14.94\pm12.49$ & $-23.42\pm10.16$ & $-7.37\pm3.19$ & $259.0\pm103.4$ \\
& $M_2$ & $0.383\pm0.287$ & $4.381\pm1.844$ & $+573.87\pm455.39$ & $+371.17\pm411.36$ & $+9.02\pm13.16$ & $79.4\pm6.1$ \\
& $M_3$ & $0.950\pm0.054$ & $1.877\pm0.427$ & $-18.20\pm19.42$ & $-25.84\pm8.58$ & $\mathbf{-9.19\pm2.05}$ & $451.3\pm99.5$ \\
& $M_4$ & $0.957\pm0.032$ & $1.523\pm0.119$ & $-20.29\pm23.32$ & $\mathbf{-27.19\pm20.52}$ & $-6.88\pm3.14$ & $457.9\pm41.8$ \\
& $M_5$ & $\mathbf{0.960\pm0.030}$ & $\mathbf{1.514\pm0.238}$ & $\mathbf{-22.29\pm19.55}$ & $-24.69\pm19.40$ & $-7.14\pm4.33$ & $\mathbf{79.0\pm7.7}$ \\
\bottomrule
\end{tabular*}
\end{table*}

\section{Structured TMIR Supervision for Imitation Learning}
For downstream validation, we evaluate the learnability and execution characteristics of the complete CMDIR supervision interface, without introducing a new imitation-learning architecture or attributing outcomes to individual interface components.

We execute each CMDIR-retargeted controller in the task environment and record synchronized images, robot states, control metrics, and applied commands. At each rollout timestamp $t_n$, the corresponding query of $R^\star$ and execution records yield $a_n^{\rm TMIR}=[(u_i,k_i,d_i)_{i\in\{W,E,S\}},p_{\rm cmd}^{\rm ref},r_{\rm cmd}^{\rm ref},g_{\rm grip}]\in\mathbb R^{31}$. The six command-pose coordinates are position and rotation vector relative to a fixed task reference pose $T_{\rm ref}$, independent of the impedance axes. Labels decode under the rollout metric to the recorded $K^\star,D^\star,T_{\rm cmd}^\star$, pairing executed controllers with the observations they produce.

ACT/TRACT policies \cite{zhao2023act,liu2026tract} predict future command chunks from currently available images and proprioception. We freeze the pretrained Cartesian policy's visual encoder and retrain the remaining modules with adapted action interfaces.

We supervise manifold parameters, decoded controllers, and control responses. Manifold errors compare scaled $k,d$ and the projective distance $1-(\hat u_i^\mathsf T\Lambda_cu_i^\star)^2$, with both axes normalized under the target-time metric. Decoded errors compare $K,D$ in Frobenius norm and command poses by position and rotation angle. Response errors compare the six components of $F_{\rm ctrl}=K(T_{\rm cmd}\ominus T_{\rm obs})-D\,v_{\rm obs}$. Each quantity is scaled using training data.

Training uses recorded states and metrics at each target timestamp; deployment uses current measurements. These losses apply to chunk queries, overlapping predictions for the same target time, and MetricReg execution:
\begingroup\small\cmdirDisplaySpacing
\begin{equation}
\begin{aligned}
\mathcal L={}&\mathcal L_{\rm org}+\sum_{c,s}\lambda_{c,s}\mathcal L_{c,s}+\lambda_g\mathcal L_{\rm grip}+\lambda_v\mathcal L_{\rm valid}.
\end{aligned}
\label{eq:tmir-loss}
\end{equation}
\endgroup
Here $c\in\{\mathrm{resp},\mathrm{dec},\mathrm{mani}\}$ and $s\in\{\mathrm{chunk},\mathrm{overlap},\mathrm{exec}\}$ define nine comparisons. Chunk and execution use recorded rollout targets; overlap compares predictions. Gripper errors use the same three locations, and $\mathcal L_{\rm valid}$ penalizes scalar repairs. $\mathcal L_{\rm org}$ retains native auxiliary losses, such as KL or phase-boundary supervision. Weights are fixed before training.

\begin{figure}[!ht]
  \centering
  \includegraphics[width=\columnwidth]{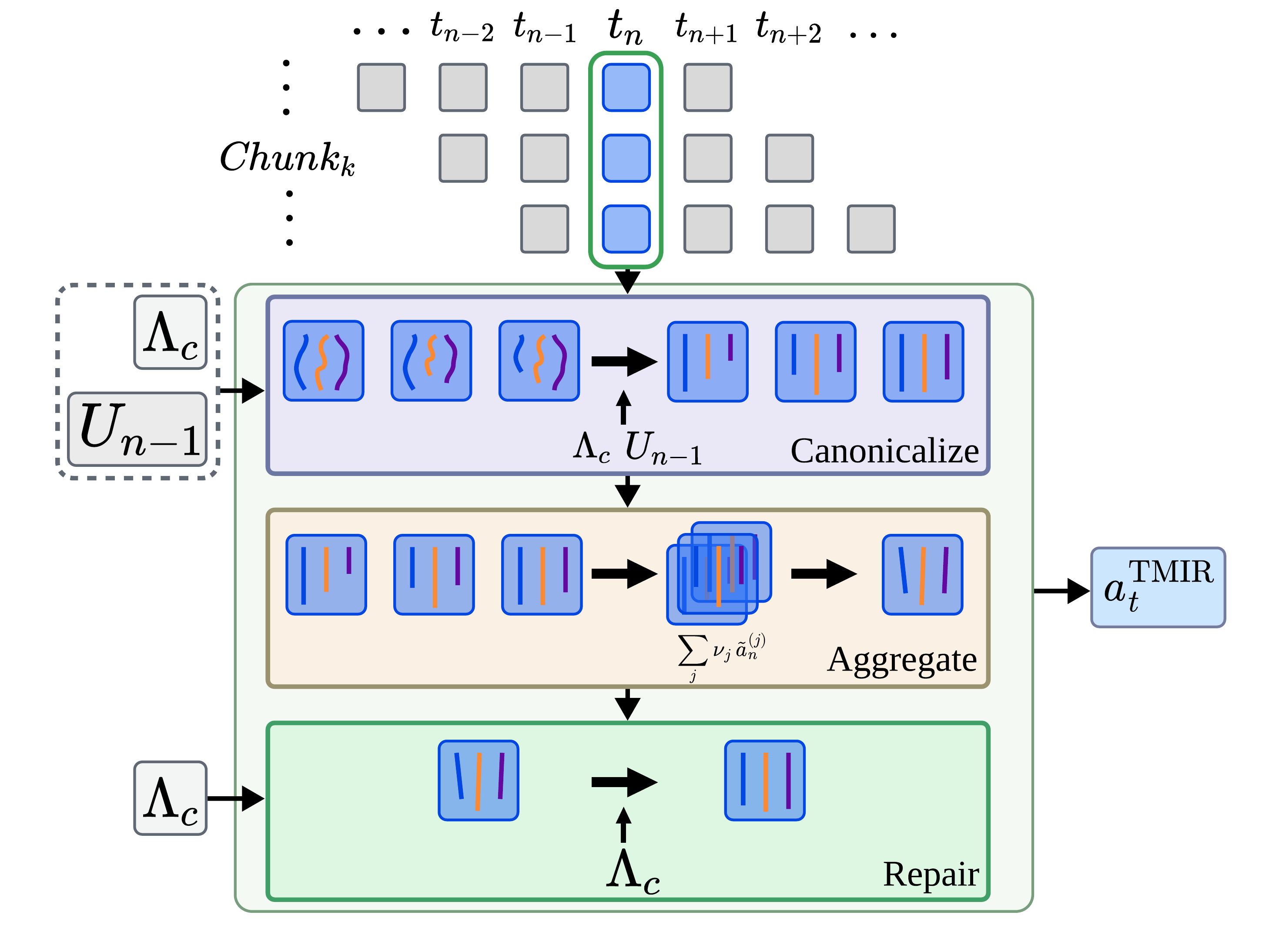}
  \caption{MetricReg canonicalizes, aggregates, and repairs overlapping predictions into one executable structured action.}
  \label{fig:metricreg}
\end{figure}

MetricReg is a deterministic execution map that normalizes axes under $\Lambda_c(t_n)$, aligns their signs to the previous executed frame, and combines overlapping predictions for the same execution time using normalized age weights $\nu_{j,n}$:
\begingroup\small\cmdirDisplaySpacing
\begin{equation}
U_n^{\rm exec}=\operatorname{Orth}_{\Lambda_c(t_n)}
\left(\sum_j\nu_{j,n}\widetilde U_n^{(j)}\right).
\label{eq:metricreg}
\end{equation}
\endgroup
Here $\widetilde U_n^{(j)}$ contains aligned axes. The same weights average $k,d,g_{\rm grip}$ and fixed-reference pose coordinates, which remain unchanged by axis sign flips. Ordered $W,E,S$ orthogonalization retains all roles; previous and coordinate axes resolve degeneracy. Scalar repair enforces the passive stiffness floor, positive damping, and gripper limits (Fig.~\ref{fig:metricreg}). Training shares this map: both overlapping predictions receive gradients, while rollout targets and the previous-frame alignment reference are detached.

\section{Experiments}
\subsection{Controller Retargeting: Overall Gains and Their Sources}
\begin{figure}[!ht]
  \centering
  \includegraphics[width=\columnwidth]{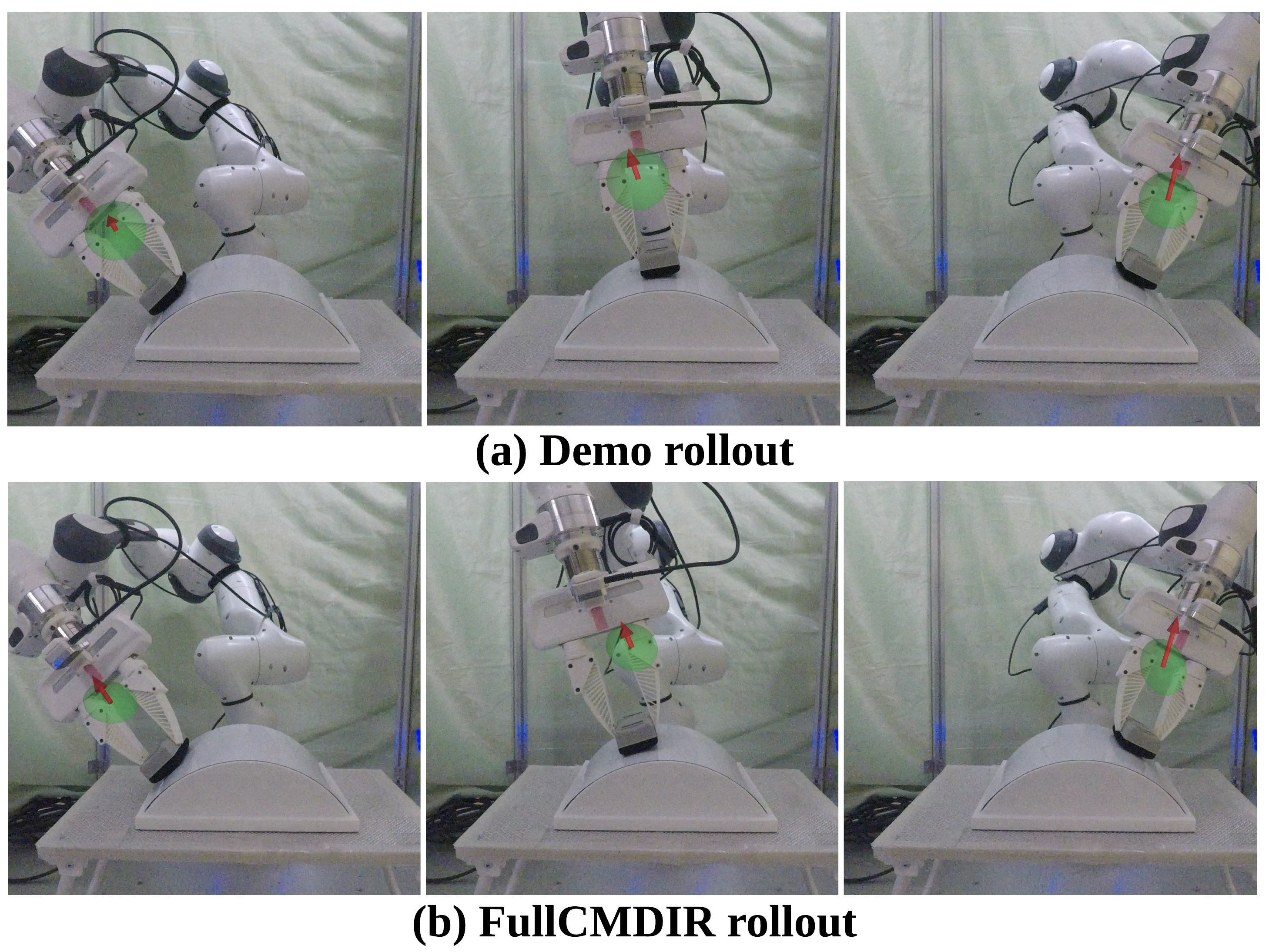}
  \caption{Representative curved-wiping rollout at matched task progress. Top: source demonstration; bottom: Full CMDIR. Red arrows denote measured contact force and green ellipsoids visualize the local impedance geometry.}
  \label{fig:controller-results}
\end{figure}

Building on MDIR's comparison with uniform gain scaling \cite{liu2026mdir}, we evaluate CMDIR on real planar wiping, curved wiping, and box pushing. Five source demonstrations per task and five variants give 75 source--method cases. Each was executed three times, totaling 225 trials. $M_1$ is discrete MDIR; $M_2$ lifts its legacy formulation into Continuous TMIR with C-MPO; $M_3/M_4$ use C-MPO with Ideal/Compromise dynamics; $M_5$ uses Compromise dynamics and FastMPO. Each source Demo supplies the paired reference.

The FastMPO structure was compiled from three development demonstrations---one curved-wiping, one planar-wiping, and one pushing instance---disjoint from all 15 Table~\ref{tab:controller} sources. The compiled structure was frozen before evaluation. Solve times cover optimization and candidate confirmation from prepared initialization; structure compilation, source preparation, and export are excluded.

Table~\ref{tab:controller} reports source-level means and standard deviations after averaging repeated executions. Wiping uses $\mathcal S_{\rm wipe}=\sqrt{\mathcal S_{\rm occ}\mathcal S_{F_z}}$: cosine similarities of spatial coverage and square-root-coverage-weighted normal-force maps, respectively. $F_z$ denotes the local surface-normal force; planar maps remove the constant placement offset. Pushing uses object displacement relative to Demo. Pose $\Delta$ is mean SE(3) deviation, converting rotation at $0.05$\,m/rad, divided by Demo's translational path length. Force fluctuation averages within-window standard deviations of bias-corrected force magnitude over Stable; peak force is its Stable maximum. Nominal power averages $|f_{\rm ctrl}^{\mathsf T}v_{\rm real}|$ over the full rollout, using controller force and measured velocity. Force and power changes are relative to Demo.

Full CMDIR ($M_5$) improves mean task retention and moderation together. Relative to $M_1$, task proxy rises from $0.696$ to $0.847$ in curved wiping, $0.796$ to $0.835$ in planar wiping, and $0.929$ to $0.960$ in pushing. Curved-wiping pose deviation falls from $0.617\%$ to $0.245\%$, while force fluctuation and peak-force changes reach $-25.74\%$ and $-28.05\%$ relative to Demo. Planar wiping and pushing show the same direction of mean improvement. Nominal power remains below Demo across tasks, with further reductions over $M_1$ concentrated in wiping. Fig.~\ref{fig:controller-results} illustrates matched-progress execution.

Direct continuous lifting ($M_2$) reduces task proxy to $0.378$, $0.472$, and $0.383$, increasing pose deviation and force fluctuation across tasks. Lower wiping power accompanies lost task effect. Each spline coefficient affects an interval and overlapping response windows, unlike pointwise optimization. This temporal coupling may explain why legacy criteria fail under direct transfer; $M_2$ does not test post-hoc interpolation.

$M_3\!\rightarrow\!M_4$ compares dynamics under C-MPO. Curved-wiping task proxy improves from $0.775$ to $0.869$ and pose deviation falls from $0.320\%$ to $0.235\%$; planar wiping and pushing also improve on both means. This supports retaining projected physical inertia and moving-frame transport for task preservation. Interaction remains task-dependent: Ideal has lower mean fluctuation and peak force in planar wiping. Across the three tasks, Compromise improves mean task retention while maintaining moderated interaction.

With FastMPO ($M_4\!\rightarrow\!M_5$), solve times fall by $7.81\times$, $9.36\times$, and $5.80\times$ in curved wiping, planar wiping, and pushing. Mean task-proxy differences stay within $0.022$, with comparable pose and interaction outcomes. Force-change variability remains large: some task-preserving samples gain little moderation.

\FloatBarrier

\subsection{Downstream Feasibility and Interaction Performance}

\begin{figure}[!ht]
  \centering
  \input{figures/Fig.ilforce}
  \par\vspace{5pt}
  \begin{minipage}[t]{0.48\textwidth}
\refstepcounter{table}
\label{tab:il}
\centering
{\footnotesize\textsc{Table~\thetable}\\[-1pt]
Real and simulated wiping: downstream learnability and successful-rollout force statistics.\par}
\vspace{4pt}
\scriptsize
\setlength{\tabcolsep}{2pt}
\begin{tabular}{@{}llccc@{}}
\toprule
Task & Supervision & \shortstack{Successful rollouts} & \shortstack{Force fluctuation (N)} & \shortstack{Peak force (N)} \\
\midrule
\multirow{2}{*}{\shortstack{\textit{Real}: Planar Wiping}}
& Ordinary & \textbf{9/10} & 0.34 $\pm$ 0.11& 8.55 $\pm$ 3.92\\
& CMDIR & 6/10 & \textbf{0.29 $\pm$ 0.13}& \textbf{5.58 $\pm$ 1.45}\\
\midrule
\multirow{2}{*}{\shortstack{\textit{Sim}: Curved Wiping}}
& Ordinary & \textbf{27/30 }& 0.29 $\pm$ 0.15& 11.64 $\pm$ 6.15\\
& CMDIR & \textbf{29/30} & \textbf{0.20 $\pm$ 0.03} & \textbf{7.83 $\pm$ 2.40}\\
\bottomrule
\end{tabular}
\end{minipage}

\end{figure}

We test whether action-chunking policies can learn the complete TMIR supervision interface and characterize the interaction behavior of successful executions. Ordinary Cartesian supervision supplies a fixed-impedance reference; CMDIR uses MetricReg. Real planar wiping uses TRACT for pickup, wiping, and return, combining positioning with force regulation. Simulated curved wiping uses ACT for changing task geometry.

Conditions share observations, action horizon, optimizer, training duration, data split, and evaluation protocol within each task. Real success requires pickup, valid wiping, and return within a predefined position tolerance; simulated success requires completing the prescribed wipe with valid contact. Force metrics cover successful, valid-contact wiping only; success rates include all trials. Fig.~\ref{fig:forcecurves} shows real wiping forces.

In Table~\ref{tab:il}, CMDIR completes 29/30 simulated and 6/10 real trials, versus 27/30 and 9/10 under ordinary supervision. Among successful valid-contact executions, CMDIR shows lower mean force fluctuation and peak force in both settings; these results establish learnability, not policy superiority or component-level causality.

\FloatBarrier

\section{Discussion and Conclusion}
CMDIR pursues task-aware compliance: retaining essential task responses while moderating avoidable interaction. It extends MDIR to continuous TMIR fields with Compromise dynamics and an offline-compiled FastMPO structure. These fields also supply structured supervision for imitation.

Across 225 trials, Compromise improves mean task retention over Ideal, and FastMPO preserves comparable outcomes with $5.8$--$9.4\times$ speedup. Relative to discrete MDIR, outcomes shift toward gentler controllers with better task retention; direct legacy lifting degrades both.

The downstream study establishes learnability, not policy superiority: successful CMDIR-supervised executions show lower force fluctuation and peak force, but real-robot completion is lower. Because rollout data, TMIR actions, MetricReg, auxiliary losses, and variable-impedance execution change jointly, no single component is credited for the force difference.

Moderation varies across demonstrations: some task-preserving executions show little force improvement, and lower forces can change force-field similarity. No uniformly preferable impedance schedule is established. Retargeting assumes matched environments; changes in surface placement, friction, or tool contact can alter responses, especially in curved wiping. First-order dynamics and finite-resolution evaluation cover modeled changes; higher-order effects and unseen environments require empirical evaluation. Query resolution remains source-limited.

Future work will learn shared task-frame patterns $\mathcal B$ and instruction patterns $\Theta$ across demonstrations, develop a TMIR-oriented imitation backbone, and investigate generalization and precision manipulation.

\bibliographystyle{IEEEtran}
\bibliography{references/references}

\end{document}